# A Complete Workflow for Development of Bangla OCR

Farjana Yeasmin Omee
Dept. of Computer Science & Engineering
Shahjalal University of Science & Technology, Sylhet

Shiam Shabbir Himel
Dept. of Computer Science & Engineering
Shahjalal University of Science & Technology, Sylhet

Md. Abu Naser Bikas
Lecturer, Dept. of Computer Science & Engineering
Shahjalal University of Science & Technology, Sylhet

## ABSTRACT

Developing a Bangla OCR requires bunch of algorithm and methods. There were many effort went on for developing a Bangla OCR. But all of them failed to provide an error free Bangla OCR. Each of them has some lacking. We discussed about the problem scope of currently existing Bangla OCR's. In this paper, we present the basic steps required for developing a Bangla OCR and a complete workflow for development of a Bangla OCR with mentioning all the possible algorithms required.

## 1. INTRODUCTION

OCR means Optical Character Recognition, which is the mechanical or electronic transformation of scanned images of handwritten, typewritten or printed text into machine-encoded text. It is widely used for converting books and documents into electronic files, to computerize a record-keeping system in an office or to publish the text on a website.

OCR has emerged a major research field for its usefulness since 1950. All over the world there are many widely spoken languages like English, Chinese, Arabic, Japanese, and Bangla etc. Bangla is ranked 5th as speaking language in the world. With the digitization of every field, it is now necessary to digitized huge volume of old Bangla book by using an efficient Bangla OCR. However, till today there is no such OCR for Bangla is developed. The existing Bangla OCRs could not fulfill the desired result. From 80's Bangla OCR development has started and now for its necessity this field becomes a major research area today. For further progression, synchronization of the total system is required. Here we present a total overview of Bangla OCR and its existing challenge is to know the development procedure and to estimate what more requires to do to create a complete Bangla OCR. Related works on Bangla OCR are given in section 2. Section 3 gives details of bangle font properties. Section 4 discusses the common steps of Bangla OCR. In

Section 5 we discussed about Tesseract OCR engine and problem scope of existing Bangla OCR is discussed in section 7. We concluded our paper in section 8.

## 2. RELATED WORKS

Though Bangla OCR is not a recent work, but there are very few mentionable works in this field. 'BOCRA and Apona-Pathak' was made publicly in 2006 [1]. But they are not open source. The Center for Research on Bangla Language Processing (CRBLP) released *BanglaOCR* – the first open source OCR software for Bangla – in 2007 [2]. *BanglaOCR* is a complete OCR framework, and has a recognition rate of up to 98% (in limited domains) but it also have many limitations.

## 3. PROPERTIES OF BANGLA FONT

Basic Bangla script consists of 11 vowels (Shoroborno), 39 consonant (Benjonborno) and 10 numerical. The basic characters are shown in Table: 1

**Table 1: Bangla basic characters**

| VOWELS | অ, আ, ই, ঈ, উ, ঊ, ঋ, এ, ঐ, ও, ঔ |
|---|---|
| CONSONENTS | ক, খ, গ, ঘ, ঙ, চ, ছ, জ, ঝ, ঞ, ট, ঠ, ড, ঢ, ণ, ত, থ, দ, ধ, ন, প, ফ, ব, ভ, ম, য, র, ল, শ, ষ, স, হ, ড়, ঢ়, য়, ৎ, ং, ঃ, ঁ |
| NUMERICAL | ০, ১, ২, ৩, ৪, ৫, ৬, ৭, ৮, ৯ |

Some common properties in Bangla language are given below:

- Writing style of Bangla is from left to right.
- As English language there are no upper and lower case in Bangla language.
- Most of the word, the vowels takes modified shape called modifiers or allograph [3-5]. Table: 2 show some example.
- There are approximately 253 compound characters composed of 2, 3 or 4 consonants [6]. Table 3: shows some example.
- All Bangla alphabets and symbols have a horizontal line at the upper part called "Matra" except এ, ঐ, ও, ঔ, খ, গ, ঙ, এঃ, থ etc.

**Table 2: Bangla modifiers example**

| Vowel | Vowel modifier | Consonant | Consonant Modifier |
|-------|----------------|-----------|--------------------|
| আ | া | য | ঃ |
| ই | ি | র | a |
| ঔ | ৌ | | |

**Table 3: compound characters example**

| Compound character | Formation of compound character |
|--------------------|---------------------------------|
| ণ্ড | ন + ড |
| ক্ট | ক + ট |
| জ্ঞ | জ + ঞ |

- The "Matra" of a character remain connected with another character which also has "Matra" in a word. Those who have no "Matra" remain isolated in a word.
- Bangla characters contain three zones which are upper zone, middle zone and lower zone [6].

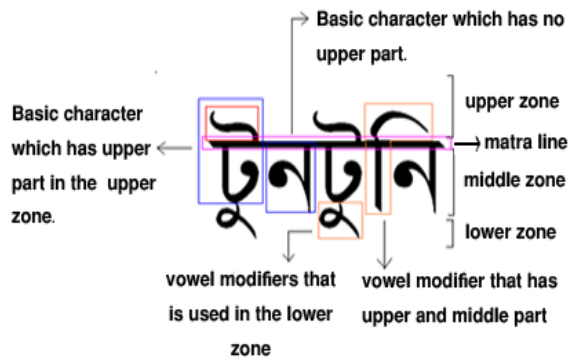

**Figure 1: Dissection of Bangla word.**

- Some characters including some modifiers, punctuations etc. have vertical stroke. [5].
- Most of the characters of Bangla alphabet set have the property of intersection of two lines in different position. Many characters have one or more corner or sharp angle property. Some characters carry isolated dot along with them [7]. i.e.: ন, ভ, ঢ়, ষ etc.

- Figure 1 summarizes an example of a Bangla Charcter with zone and modifiers.

From all of these properties, there are approximately 310 types of pattern to be recognized for Bangla script.

# 4. STEPS OF BANGLA OCR

Although OCR system can be develop for different purposes, for different languages, an OCR system contains some basic steps.

Figure-2 describes the basic steps of an OCR. A basic OCR system has the following particular processing steps [8]:

1. Scanning.
2. Preprocessing.
3. Feature extraction or pattern recognition.
4. Recognition using one or more classifier.
5. Contextual verification or post processing

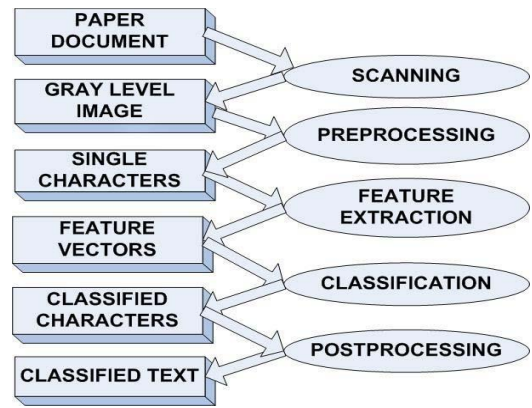

**Figure 2: Basic Steps of an OCR**

## 4.1 Scanning:

To extract characters from scaned images it is necessary to convert the image into proper digital image. This process is called text digitization. The process of text digitization can be performed either by a Flat-bed scanner or a hand-held scanner. Hand held scanner typically has a low resolution range. Appropriate resolution level typically 300-1000 dots per inch for better accuracy of text extraction [8].

## 4.2 Preprocessing:

Preprocessing consists of number of preliminary processing steps to make the raw data usable for the recognizer. The typical preprocessing steps included the following process:

4.2.1 Binarization
4.2.2 Noise Detection & Reduction
4.2.3 Skew detection & correction
4.2.4 Page layout analysis
4.2.5 Segmentation

### 4.2.1 Binarization methods

Binarization is a technique by which the gray scale images are converted to binary images. Some binarization methods are given below:

- ***Global Fixed Threshold:*** The algorithm chooses a fixed intensity threshold value I. If the intensity value of any pixel of an input is more than I, the pixel is set to white otherwise it is black. If the source is a color image, it first has to be converted to grey level using the standard conversion [9].

- ***Otsu Global Algorithm:*** This method is both simple and effective. The algorithm assumes that the image to be threshold contains two classes of pixels and calculates the optimum threshold separating those two classes so that their combined spread (intra-class variation) is minimal [10].

- ***Niblack's Algorithm:*** Niblack's algorithm calculates a pixel wise threshold by sliding a rectangular window over the grey level image. The threshold is computed by using the mean and standard deviation, of all the pixels in the window [9].

- ***Adaptive Niblack's Algorithm:*** In archive document processing, it is difficult to identify suitable sliding window size SW and constant k values for all images, as the character size of both frame and stroke may vary image by image. Improper choice of SW and k values results in poor binarization. Modified Niblack's algorithm allows automatically chosen values for k and SW, which is called adaptive Niblack's algorithm [9].

- ***Sauvola's Algorithm:*** Sauvola's algorithm is a modification of Niblack's which is claimed to give improved performance on documents in which the background contains light texture, big variations and uneven illumination. In this algorithm, a threshold is computed with the dynamic range of the standard deviation [9].

### 4.2.2 Noise detection & correction methods

Noise can be produced during the scanning section automatically. Two types of noises are common. They are background noise and salt & paper noise. Complex script like Bangla, we cannot eliminate wide pixels from upper or lower portion of a character because it may not only eliminate noise but also the difference between two characters like অ and আ and for some other characters [11]. Some noise detection methods are given below:

- The commonly used approach is, to low pass filter the image and to use it for later processing. The objective in the design of a filter to reduce noise is that it should remove as much of the noise as possible while retaining the entire signal [8].

- Mixed noise of Gaussian and impulse cannot be removed by conventional method at the same time. In [12], the author proposed an enhanced TV (Total Variation) filter which can remove these two types of noise based on PSNR (Pick Signal to Noise Ratio) and subjective image quality. The main advantage of this method is it can eliminate mixed noise efficiently & quickly.

- Comparing with the conventional method such as averaging method and the median method, the proposed

method in [13] gets higher quality. This method is based on kalman filter.

- Dots existing in a character like অ may be treated as noise. In [14], the author proposed a new method for removing noises similar to dots from printed documents. In this method, first estimate the size of the dots in each region of the text. Then the minimum size of dots in each region is estimated based on the estimated size of dot in that region.

- In [1], the authors used connected component information and eliminated the noise using statistical analysis for background noise removal. For other type of noise removal and smoothing they used wiener and median filters [15]. Connected component information is found using boundary finding method (such as edge detection technique). Pixels are sampled only where the boundary probability is high. This method requires elaboration in the case where the characteristics change along the boundary.

- In [16], noise is removed from character images. Noise removal includes removal of single pixel component and removal of stair case effect after scaling. However, this does not consider background noise and salt and paper noise.

### 4.2.3 Skew detection & correction methods

Sometimes digitized image may be skewed and for this situation skew correction is necessary to make text lines horizontal. Skew correction can be achieved in two steps. First, estimate the skew angle θt and second, rotate the image by θt, in the opposite direction [8]. Here detect the skew angle is using Matra. Two types of methods discussed in [17] & [18] for this purpose. Hough transform technique may be applied on the upper envelopes for skew estimation, but this is slow process. So in [17] an approach is discussed which is fast, accurate and robust. The idea is based on the detection of DSL segments from the upper envelope. In [18], they applied Radon transform to the upper envelope to get the skew angle. Radon transform and the Hough transform are related but not the same. Then applied generic rotation algorithm for skew correction and then applied bi-cubic interpolation.

### 4.2.4 Page Layout analysis

The overall goal of layout analysis is to take the raw input image and divide it into non-text regions and "text lines"–sub images of the original page image that each contains a linear arrangement of symbols in the target language. Some page layout algorithms are: RLSA [19], RAST [20] etc. Existing page layout analyzers are described in [21]. An algorithmic approach is described in [22] with its limitation. Still this section is developing and thus become a major research field. Using page layout analysis in OCR helps to recognize text from newspaper, magazine, documents etc.

### 4.2.5 Segmentation methods

This is the most vital and important portion for designing an efficient Bangla OCR because the next steps depends on this phase to make those process successful. Basic steps of segmentation are as follow:

- **Line Segmentation Process:** The lines of a text block are detected by scanning the input image horizontally. Frequency of black pixels in each row is counted in order to construct the row histogram. When frequency of black pixels in a row is zero it denotes a boundary between two white pixels consecutive lines.

- **Word Segmentation Process:** When a line has been detected, then each line is scanned vertically for word segmentation. Number of black pixels in each column is calculated to construct column histogram. When no black pixel is found in vertical scan that is considered as the space between two words. In this way we can separate the words. Figure-3 shows how the word is segmented.

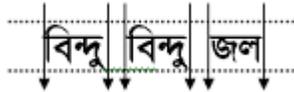

**Figure 3: Word Segmentation**

During this process, a typical situations occur when matraless character is used in a word [23].

- **Character Segmentation Process:** It is the most challenging part to build printed Bangla OCR. As Bangla is an inflectional language, the ornamentation of the characters in a word makes the segmentation difficult. A word can be constructed by:
    - Only taking the basic characters like vowels, consonants and numerals
    - Basic characters and consonant modifiers
    - Basic characters and vowel modifiers
    - Compound characters
    - Compound characters and consonant modifiers
    - Compound characters and vowel modifiers

To segmented character some process is follows here.

- **Matra line detection process:** To segment the individual character from the segmented word, we first need to find out the headline of the word which is called 'Matra'. Matra is a distinct feature in Bangla. The row with the highest frequency of black pixels is detected as matraline or headline. For the larger font size, it is observed that the height or thickness of the matraline increases. In order to detect the matraline with its full height, the rows with those frequencies are also treated as matraline [16].

- **Base line detection process:** A baseline is an imaginary line. It is a row from where the middle zone ends and lower zone starts. Baseline and end row of the line is same when the line does not have any lower modifier(s). In a general document, it is observed that about 70% lines have lower modifiers, 30% lines do not have lower modifiers. So baseline detection is very important for Bangla. The process is described in [23]. After determining the baseline, a depth first search (DFS) easily extracts the characters below the baseline[8].

- **Character detection:** The main parts of all the characters are placed in the middle zone. So the middle zone area is considered as the character segmentation portion. Since matra line connects the characters together to form a word, it is ignored during the character segmentation process to get them topologically disconnected [24]. A word constructed with basic characters is segmented into characters in a way by scanning vertically [23]. In this technique, the two characters, আ and আ, get split into two pieces due to ignoring matra line. This problem is overcome by joining the left piece to the right one to make an individual character by considering the fact that a character in the middle zone always touches the baseline [24]. There are four kinds of modifiers based on their uses. One kind of modifiers, used only in the middle zone, is called middle zone modifiers, for example ি, ী, ে and the modifiers which are used only in the lower zone, such as, ূ, ু, ্ are called the lower zone modifiers. Another kind of modifiers is there which consists of both upper and middle zone. They are like ৌ, ৈ, ৗ, ৗ. The last kind of modifier is called the upper zone modifier, such as . The character segmentation process including modifiers has been described in [16], [23], [24], [25].

## 4.3 Feature extraction or Pattern recognition

**Feature extraction** or Intelligent Character Recognition (ICR), is a much more sophisticated way of spotting characters. Feature extraction is an OCR method for classifying characters which nowadays is being used by most OCR programs. With feature extraction all characters will be divided into geometric elements like lines, arcs and circles and the combination of these elements will be compared with stored combinations of known characters. This method provides much more flexibility than the formerly used pattern recognition and also copes well with variations in font style and size.

Most modern omnifont OCR programs work by feature detection rather than pattern recognition. Some use neural networks. Various steps of feature extraction are described in [8]. Some common feature extraction methods for Bangla script are given in [16], [26], [24].

**Pattern recognition** is an OCR method for classifying characters which because of its lack of flexibility is rarely used anymore. With pattern recognition the scanned characters will be compared with stored patterns after segmentation and assigned to the pattern they match most closely. This method from the early days of OCR is suited only for the recognition of fixed fonts and is used nowadays only for a few special applications. In contrast to pattern recognition, pattern matching is generally not considered a type of machine learning, although pattern-matching algorithms can sometimes succeed in providing similar-quality output to the sort provided by pattern-recognition algorithms. So this is a static way to develop an OCR.

## 4.4. Classification

This is the last phase of the whole recognition process. Several approaches have been used to identify a character based on the features extracted using algorithms described in previous section. However, there is no benchmark databases of character sets to test the performance of any algorithm developed for Bengali character recognition [27]. Each paper has taken their own samples for training and testing their proposals. In choosing classification algorithms, use of Artificial Neural Network (ANN) is a popular practice because it works better when input data is affected with noise. Some classifiers are

- Decision tree [24].
- MLP [28].
- Kohonen Neural Network [29]
- HMM(Hidden Markov Model)[30]

### 4.5. Post Processing

After recognition process, remain step is post processing. Sometimes the recognized character does not match with the original one or cannot be recognized from the original. To check this post processing steps include spell checking, error checking, and text editing etc. process.

## 5. TESSERACT OCR ENGINE

The open Source OCR landscape got dramatically better recently when Google released the Tesseract OCR engine as open source software. Tesseract is considered one of the most accurate free software OCR engines currently available. Currently research & development of OCR lay upon the Tesseract. Tesseract is a powerful OCR engine which can reduce steps like feature extraction and classifiers. The Tesseract OCR engine was one of the top 3 engines in the 1995 UNLV Accuracy Test. Between1995 and 2006 however, there was very little activity in Tesseract, until it was open-sourced by HP and UNLV in 2005; it was again re-released to the open source community in August of 2006 by Google. A complete overview of Tesseract OCR engine can be found in [31]. The complete source code and different tools to prepare training data as well as testing performance is available at [32]. The algorithms used in the different stages of the Tesseract OCR engine are specific to the English alphabet, which makes it easier to support scripts that are structurally similar by simply training the character set of the new scripts. There are published guidelines about the procedure to integrate Bangla/Bengali language recognition using Tesseract OCR engine [33]. Detailed knowledge of the Bangla script as well as the character training procedure is also given in [33].

## 6. PROBLEM SCOPE IN BANGLA OCR

Each author has used their own set of data. As a result, comparative analysis does not produce a really meaningful result. Some authors addressed noise detection and cleaning phase in their works. However, a comprehensive solution for elimination of all types of noise is not available. The reader has already understood that Bangla has not only basic characters; it is rich with modifiers and compound characters. Placement of modifiers may happen on the upper, lower, left or right side of original characters which generates a lot of complications. Rarely authors could confidently claim that a particular segmentation and classification scheme has dealt with all of them. Again lack of standard or benchmark samples do not allow one to make a comprehensive testing of their application. Existing applications are build up for printed documents where complex structure of documents is assumed not to present. Also these applications do not have any good quality page layout analyzer which could automatically identify picture and text paragraph of an input image. These have some limitations on line segmentation for newspaper document images because of the appearance of joining problem between two lines in the image. For medium quality printed low resolution document images, existing OCR does not perform well.

## 7. CONCLUSION

In this paper, we described the whole Bangla OCR process step by step mentioning the algorithms required for each step. We also covered the flaws of already developed Bangla OCR. Total workflow of OCR is summarized in this paper so that it will be helpful for the developers and researchers on the research field of OCR. We are currently working on developing an error free and full functional Bangla OCR.